\documentclass[cameraready]{Interspeech}

\usepackage{multirow}
\usepackage{amsmath}
\usepackage{algorithm}
\usepackage{algorithmic}
\title{Towards Efficient Simultaneous Inverse Text Normalization with Pretrained Text-to-Text Language Model and Read–Tag–Write Policy}

\author[affiliation={1}, orcid=0000-0001-8438-6585, equalcontribution, correspondingauthor]{Kiet Anh}{Hoang}
\author[affiliation={1}, orcid=0009-0003-1190-5569, equalcontribution]{Khanh}{Le}
\author[affiliation={1}, orcid=0009-0009-4515-7056]{Bao}{Nguyen}
\author[affiliation={1}, orcid=0009-0003-1560-2813]{Linh}{Pham}
\author[affiliation={1}, orcid=0009-0001-1973-2346]{Dung}{Vo}
\author[affiliation={1}, orcid=0009-0002-1422-9685]{\\Thai}{Tran}
\author[affiliation={1}]{Mai}{Nguyen}
\author[affiliation={1}, orcid=0009-0005-5402-3683]{Tri}{Nguyen}
\author[affiliation={1}]{Vu}{Le}

\address{
    $^1$ UNEY, Switzerland
}

\email{kiet.ha@uney.com, khanh.ld@uney.com}


\usepackage{comment}

\begin{document}

\maketitle

\begin{abstract}

    Streaming Inverse Text Normalization (ITN) is vital for converting spoken-form outputs from streaming Automatic Speech Recognition into formatted written text. Existing streaming ITN methods rely on hybrid systems combining neural tagging with expert-crafted finite-state transducer rules, limiting scalability across domains and languages. While end-to-end models offer superior scalability by learning directly from data, standard encoder-decoder architectures are inherently non-streaming due to global attention. In this paper, we propose an efficient streaming end-to-end ITN system, adapting a pretrained text-to-text model to leverage its robust linguistic knowledge. To enable streaming, we introduce architectural adaptations, a specialized training strategy, a Read–Tag–Write decoding policy, and inference optimizations. Experiments on Vietnamese datasets show accuracy comparable to non-streaming baselines, outperforming hybrid methods while satisfying real-time latency requirements.

\end{abstract}

\section{Introduction}

Inverse Text Normalization (ITN) converts spoken-form transcriptions produced by Automatic Speech Recognition (ASR) systems into readable written-form representations~\cite{sunkara2021neural}. While ASR models typically train on verbalized labels to maintain phonetic alignment with audio, their outputs often lack readability and impair downstream tasks such as machine translation and natural language understanding. ITN addresses this by restoring punctuation, applying capitalization, and performing phrase-level transformations, including semiotic and phonetic phrases~\cite{nguyen2023adapitn}. For example, it converts expressions like ``one point two dollars'' to ``\$1.20''. While spoken and written forms are frequently identical, certain structured phrases differ substantially and require contextual reasoning to resolve ambiguities. This makes ITN inherently context-dependent and necessitates robust contextual representations to ensure accurate disambiguation across diverse linguistic patterns~\cite{gaur2023streaming}.

Recent advances in ITN have focused on end-to-end approaches, which achieve strong performance across languages and domains~\cite{sunkara2021neural} by implicitly learning patterns from data, thereby offering greater scalability and enabling effective model compression~\cite{gaur2023streaming}. 
These methods primarily use sequence-to-sequence (seq2seq) architectures to map spoken-form inputs directly to written-form outputs~\cite{sunkara2021neural, nguyen2023adapitn, ihori2020large, chen23s_interspeech, choi2024spoken}. However, standard seq2seq models often produce unrecoverable hallucination errors, yielding fluent yet semantically incorrect results that alter original meaning~\cite{sunkara2021neural}. Additionally, processing full sentences increases computational cost and risks unnecessary changes to unchanged tokens~\cite{nguyen2023adapitn}. Hybrid methods alleviate these problems by combining neural tagging models with finite-state transducer (FST) rules~\cite{shugrina2010formatting, zhang21ja_interspeech, antonova22_interspeech}. They excel at preserving unmodified text but depend on expert-crafted FST grammars and extensive inside-outside-beginning (IOB) tag annotation for each ITN category. This dependence limits scalability and performance improvement due to the substantial manual effort required for IOB labeling and FST rule creation~\cite{gaur2023streaming}. Pretrained language models~\cite{sunkara2021neural, choi2024spoken} and data-driven techniques~\cite{chen23s_interspeech, paul22_interspeech} have also been explored to address data scarcity, improve generalization, and enhance ITN performance across diverse languages and domains.

Despite advances in offline ITN, simultaneous or streaming ITN remains underexplored yet essential for real-time applications such as live ASR, meeting transcription, and speech translation, where formatted text must be displayed instantaneously to preserve conversational flow. Industrial systems often rely on cascaded architectures, applying incremental punctuation and casing models as post-processing steps~\cite{shugrina2010formatting, zhang21ja_interspeech}. While these modular pipelines provide stability, they struggle with complex semiotic ambiguities that require joint modeling of capitalization, punctuation, and structural transformations. Pioneering work in this area~\cite{gaur2023streaming} introduced a hybrid streaming solution employing a chunk-based Transformer~\cite{vaswani2017attention} tagger to detect ITN categories in incoming ASR tokens, followed by category-specific FSTs for targeted conversion. A subsequent study~\cite{ho25_interspeech} adapted a pretrained language model with dynamic, context-aware processing to handle variable ASR chunk sizes and accumulated tokens, achieving low latency and accuracy comparable to non-streaming ITN on a limited dataset. However, current streaming methods rely exclusively on hybrid architectures and inherit limited scalability due to the need for manual FST rule development and extensive IOB labeling. While seq2seq models offer greater scalability, they cannot operate in a streaming fashion because their encoder-decoder architectures rely on global cross-attention, which prevents incremental processing.

\begin{figure*}[!ht]
  \centering
  \includegraphics[width=0.98\linewidth]{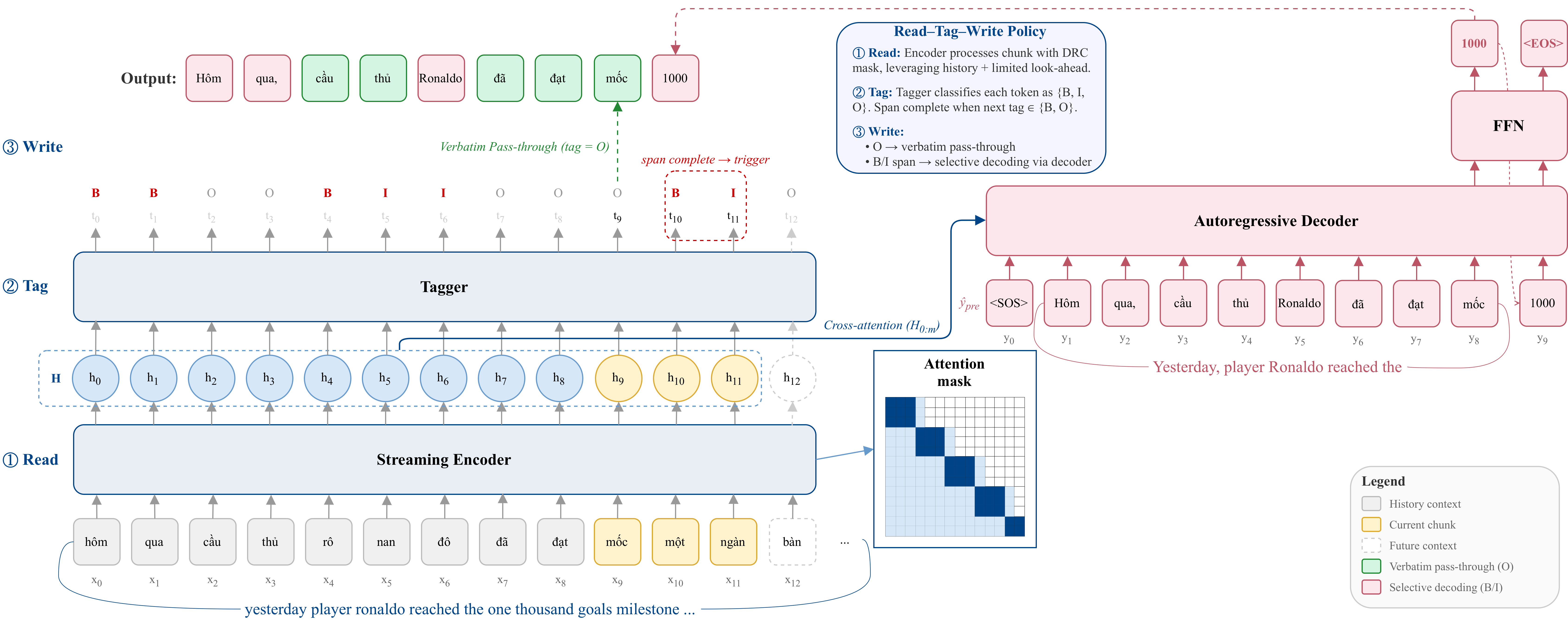}
  \caption{Overview of the proposed streaming end-to-end ITN framework using the Read–Tag–Write policy. The streaming encoder utilizes Dynamic Right Context masking to process incoming chunks of spoken-form tokens, while a joint tagger predicts \{B, I, O\} labels. Verbatim tokens (O) are passed through instantly to minimize latency. Upon the completion of a normalization span, the autoregressive decoder is selectively triggered, generating normalized text by cross-attending to the relevant encoder hidden states ($H_{0:m}$) and conditioning on the written-form prefix $\hat{y}_{pre}$.}
  \label{fig:our_model}
\end{figure*}

To adapt offline seq2seq models for streaming inference, the field of Simultaneous Machine Translation (SiMT) has introduced various techniques to enable incremental processing~\cite{wang2025conversational}. A core component of SiMT is the read-write policy, which decides whether to wait for new source tokens (\textsc{read}) or generate target tokens (\textsc{write}) to balance quality and latency. For instance, the wait-$k$ policy~\cite{ma2019stacl, elbayad20_interspeech, zhao2023adaptive} enforces a fixed lag by alternating \textsc{write}/\textsc{read} operations after an initial $k$-token delay, while monotonic attention mechanisms~\cite{arivazhagan2019monotonic} restrict attention to left-to-right alignments. Other adaptive strategies involve learned thresholds for dynamic read-write decisions~\cite{zheng2019simpler}, Hidden Markov Transformers for simultaneous generation~\cite{zhang2023hidden}, or leveraging large language models as policy-makers~\cite{guo2025large}. However, while effective for translation, these strategies introduce redundant computation for ITN.
In SiMT, the reordering problem represents a primary conflict between translation quality and real-time latency. Unlike SiMT, which requires global structural restructuring, ITN is characterized by local transformations that maintain the original word order. Because ASR outputs are largely verbatim, the fixed lags and generative overhead typical of SiMT policies may be counterproductive for ITN. Consequently, the applicability of complex SiMT scheduling to the localized domain of ITN remains unexplored.

In this paper, we propose, to the best of our knowledge, the first streaming end-to-end ITN model. Our approach achieves accuracy comparable to non-streaming end-to-end baselines, outperforms streaming hybrid methods, and eliminates the need for manual FST rules and IOB labeling. Our experiments are conducted on Vietnamese, a challenging language for ITN due to its complex morphological nuances and frequent phonetic borrowing. By relying strictly on standard text-to-text foundations rather than language-specific heuristics, our approach ensures cross-linguistic generalization and is compatible with any seq2seq architecture. Our main contributions include:
\begin{itemize}
    \item \textbf{A recipe for making pretrained seq2seq models truly streamable:} We present a practical set of techniques: architectural changes, a streaming-aware multi-task objective, and a novel data augmentation technique. That together enable end-to-end ITN with strict left-to-right processing.
    \item \textbf{An efficient Read–Tag–Write decoding policy with KV Caching and Copy–Paste Strategy:} We introduce a simultaneous inference procedure that (i) passes through non-normalized tokens with minimal delay and (ii) triggers generation only upon completing a normalization span, further accelerated by KV caching and a Copy–Paste strategy that keeps historical context and removes redundant computation.
    \item \textbf{Comprehensive empirical evaluation:} We rigorously evaluate our approach on a newly curated 5M-sentence Vietnamese dataset featuring automated alignments and boundary tags, verifying its highly efficient performance.
\end{itemize}

\section{Streaming End-to-End Inverse Text~Normalization}
\label{sec:method}

In this section, we explain the chosen pretrained text-to-text models and describe our proposed framework for streaming end-to-end ITN. 

\subsection{Pretrained Text-to-Text Models}
Pretrained text-to-text models such as BART \cite{lewis2020bart} and T5 \cite{raffel2020exploring} have advanced seq2seq learning by unifying diverse NLP tasks under a generative framework. Unlike encoder-only architectures, these models utilize a complete Transformer encoder-decoder structure where both inputs and outputs are represented as text strings. BART \cite{lewis2020bart} is pretrained using a denoising objective including token masking, deletion, and sentence permutation to learn robust bidirectional encoder representations and autoregressive decoder capabilities. Similarly, T5 \cite{raffel2020exploring} employs a span-corruption objective on the Colossal Clean Crawled Corpus (C4), facilitating effective transfer learning for downstream tasks such as translation and summarization. Extensive empirical evidence indicates that leveraging these pretrained foundations significantly improves performance on downstream tasks compared to training from scratch, especially in low-resource or domain-specific scenarios~\cite{lewis2020bart, raffel2020exploring, xue2021mt5}.

For the Vietnamese language, EnViT5\footnote{\url{https://huggingface.co/VietAI/envit5-base}} \cite{ngo2022mtet} adapts the T5-base architecture to a bilingual setting. Following the standard configuration, the model features 12 layers in both the encoder and decoder, a hidden dimension of 768, 12 attention heads, and a gated-GELU feed-forward projection with an intermediate dimension of 2048, totaling approximately 275 million parameters. EnViT5 was pretrained on a massive bilingual corpus, including an 80GB filtered subset of CC100 \cite{wenzek2020ccnet}, using a 15\% span-corruption objective for both English and Vietnamese. It utilizes a shared SentencePiece tokenizer~\cite{kudo-richardson-2018-sentencepiece} with a 50,048-subword vocabulary specifically optimized for the morphological nuances of the Vietnamese language. EnViT5 achieves superior performance on Vietnamese-English translation tasks on the PhoMT benchmark \cite{doan2021phomt}, outperforming larger models such as mBART \cite{ngo2022mtet}. Given its superior semantic alignment and generative efficiency, we adopt EnViT5 as the foundation for our end-to-end ITN system, subsequently adapting the architecture to satisfy streaming requirements.

\subsection{Model Architecture}
The proposed streaming end-to-end ITN model, illustrated in Fig.~\ref{fig:our_model}, is built upon the pretrained EnViT5 architecture. It comprises a streaming-capable encoder and an autoregressive decoder, optimized for real-time inverse text normalization.

\subsubsection{Streaming Encoder with Joint Tagger}
To support streaming inference within the encoder, we employ the Dynamic Right Context (DRC) masking technique~\cite{le24_interspeech} during training. Specifically, we utilize an attention mask to incorporate variable chunk sizes and dynamic look-ahead while maintaining access to the full historical left context. By varying these parameters during training, the masking strategy effectively simulates the restricted future context inherent in real-time processing, ensuring robustness to diverse latency constraints during inference.

For an input sequence $X = \{x_0, \dots, x_{n-1}\}$, the encoder generates hidden representations $H \in \mathbb{R}^{n \times d}$ through Transformer layers constrained by the DRC mask, where $d$ denotes the hidden dimensionality of the model. A linear tagging layer (the Tagger module as illustrated in Fig.~\ref{fig:our_model}), is integrated atop these hidden representations to facilitate on-the-fly detection of normalization spans (e.g., punctuation, capitalization, or numerical expressions). Following~\cite{nguyen2023adapitn}, the tagger classifies each token $x_i$ using a $\{\text{B, I, O}\}$ (Beginning, Inside, Outside) labeling scheme. The predicted tag $\hat{t}_i$ is determined by:

\begin{equation}
\hat{t}_i = \text{argmax}(H_i W_{tag}), \quad W_{tag} \in \mathbb{R}^{d \times 3}
\end{equation}
These predicted tags serve as the control signals for the Read–Tag–Write policy (described in Section~\ref{section:rtw}). Unlike hybrid approaches \cite{gaur2023streaming,ho25_interspeech} that require fine-grained tags for various ITN categories to trigger specific FST rules, our tagger utilizes only 3 classes. This reduced label space simplifies classification and leads to more robust span detection compared to complex multi-class systems.

\subsubsection{Autoregressive Decoder and Training Objective}
During training, the decoder generates written-form tokens $y_t$ by cross-attending to the encoder representations $H$, conditioned on the ground-truth previous tokens $y_{<t}$. While the encoder is constrained for streaming, the decoder maintains its standard autoregressive structure. The probability distribution over the vocabulary at each step is computed as:
\begin{equation}
P(y_{t}|y_{<t},H) = \text{softmax}(e_{t} W_{ffn}^{\top})
\end{equation}
where $e_t \in \mathbb{R}^{d}$ is the decoder hidden state at step $t$, $W_{ffn} \in \mathbb{R}^{V \times d}$ is the vocabulary projection matrix, and $V$ denotes the vocabulary size.

The model is optimized using a multi-task objective. We define $\mathcal{L}_{tag}$ as the cross-entropy loss for span detection and $\mathcal{L}_{gen}$ as the sequence generation loss:

\begin{equation}
\mathcal{L}_{tag} = - \frac{1}{\sum_{b=1}^{B} N_b} \sum_{b=1}^{B} \sum_{i=1}^{N_b} \log P(t_{b,i} \mid x_{b,i})
\end{equation}

\begin{equation}
\mathcal{L}_{gen} = - \frac{1}{\sum_{b=1}^{B} T_b} \sum_{b=1}^{B} \sum_{t=1}^{T_b} \log P(y_{b,t} \mid y_{b,<t}, H_b)
\end{equation}
Here, $B$ denotes the batch size, while $N_b$ and $T_b$ represent the source and target sequence lengths for the $b$-th sample, respectively. The terms $t_{b,i}$ and $y_{b,t}$ denote the ground-truth tag at step $i$ and the target token at step $t$. The probability $P(t_{b,i} \mid x_{b,i})$ is computed by applying the softmax function to the tagger logits~$H_{b,i} W_{tag}$.

The overall objective is a combination of these components:
\begin{equation}
\mathcal{L} = \mathcal{L}_{tag} + \mathcal{L}_{gen}
\end{equation}

\subsection{Read–Tag–Write Policy}
\label{section:rtw}


To enable efficient streaming inference, we propose the Read–Tag–Write (RTW) policy. Unlike SiMT policies that manage global structural reordering~\cite{wang2025conversational}, RTW exploits the local, order-preserving nature of ITN to coordinate the interaction between the streaming encoder and the autoregressive decoder, ensuring that generative normalization occurs only when necessary. The policy operates in three distinct phases:

\begin{enumerate}[leftmargin=*]

\item \textbf{Read:} The system receives incoming ASR tokens in discrete chunks. Let $C_k = \{x_i, \dots, x_{i+c-1}\}$ be the $k$-th chunk of size $c$. The encoder, constrained by the DRC mask, processes $C_k$ by leveraging the cached historical context and a limited look-ahead of $r$ tokens to compute hidden representations $H_{i:i+c-1}$, which are appended to the global hidden state sequence $H_{0:i+c-1}$.

\item \textbf{Tag:} A joint tagging layer performs token-level classification on the hidden states of the newly arrived chunk. For each token $x_j \in C_k$, the model predicts a tag $\hat{t}_j \in \{B, I, O\}$ as:
\begin{equation}
\hat{t}_j = \text{argmax}(H_j W_{tag})
\end{equation}
The policy identifies transformation spans using $\{B, I\}$ labels. A span (e.g., a single $B$ or a $B$-$I$ sequence) is considered complete only when the tagger encounters a tag $\hat{t}_{j+1} \in \{B, O\}$ or a sentence boundary. If $\hat{t}_{j+1} = I$, the token is treated as a continuation of the current span and the system continues to ``Read" without triggering the ``Write" phase.

\item \textbf{Write:} Unlike standard seq2seq models that decode the entire sequence, the RTW policy invokes the decoder selectively based on predicted tags:

\begin{itemize}
    \item \textbf{Verbatim Pass-through:} For tokens tagged as $O$, the system directly emits the spoken-form token. This ensures minimal latency and preserves the integrity of verbatim segments by bypassing the generative decoder.

    \item \textbf{Selective Decoding:} When a $B-I$ span (or a sequence of consecutive spans) is completed, the decoder is triggered. To optimize throughput, multiple consecutive spans within a chunk are concatenated and processed in a single decoder invocation. Let $m$ be the index of the final token in the last completed span. The decoder generates the written-form segment $\hat{y}_{local}$ by cross-attending to the encoder hidden states $H_{0:m}$:
    \begin{equation}
    P(\hat{y}_{local} \mid \hat{y}_{pre}, H_{0:m}) = \prod_{u=1}^U P(y_u \mid [\hat{y}_{pre} ; y_{<u}], H_{0:m})
    \label{eq:selective_decoding}
    \end{equation}
    where $[\cdot ; \cdot]$ denotes the concatenation operator, $\hat{y}_{pre}$ represents the written-form prefix generated in previous steps, and $y_u$ is the $u$-th token of the local sequence of length~$U$. Crucially, cross-attention is restricted to $H_{0:m}$, ensuring look-ahead context does not influence the current transformation. This constraint prevents the decoder from over-generating or emitting extraneous tokens beyond the target span. The decoder generates tokens until an \texttt{<EOS>} token is predicted, signaling the end of the local transformation. To prevent infinite decoding loops in edge cases where the model fails to emit an \texttt{<EOS>} token, we implement a forced hard-stop that truncates generation at a maximum of $L_{\text{max\_local}}$ tokens per local span.
\end{itemize}

\end{enumerate}

\subsection{Inference Optimization via Copy–Paste Strategy and KV Caching}
\label{sec:infer_kv_cache}

\textbf{Encoder KV Caching:} To maximize efficiency, we implement an incremental Key-Value (KV) cache for the encoder, enabling the reuse of computations from previous chunks. Similarly, we cache the encoder hidden states and the previously decoded written-form prefix. When the decoder is triggered, the entire cross-attention and self-attention KV states need to be updated to incorporate the newly computed encoder states. We perform an initial single batched forward pass through the decoder to populate the self-attention and cross-attention KV caches using the existing prefix, bypassing the need for autoregressive decoding during this initialization phase. Subsequent tokens are then produced via standard autoregressive generation. These optimizations significantly improve inference throughput and minimize latency. \\
\textbf{Copy–Paste Strategy:} In contrast to \cite{nguyen2023adapitn}, which completely excludes historical context from the autoregressive decoder, we argue that such context is vital for accurate normalization. Since spoken and written forms are frequently identical in ITN, the $O$ tag is the most prevalent. Our RTW policy leverages this by bypassing the decoder for these tokens and copying them directly to the output. Crucially, these skipped tokens are still incorporated into the historical context to guide the decoding of $B-I$ tokens. This approach not only reduces computational overhead and mitigates hallucinations in verbatim segments but also allows the decoder to maintain the full historical context efficiently. \\
\textbf{Decoder KV Caching:} To mitigate the computational overhead inherent in the Copy–Paste mechanism and autoregressive decoding, where standard processing would redundantly recompute the entire history for both copied and emitted tokens, we implement the ``fully streaming'' version by enabling the decoder KV Caching for both self- and cross-attention. This allows the model to reuse states from previous invocations and perform updates only for new tokens. While this eliminates initial passes, it introduces
a training-inference mismatch: training uses the full target prefix, whereas streaming inference updates states incrementally without re-attending to past tokens under new encoder states. Our experiments show this results in only a marginal performance fluctuation compared to the first strategy. 

\subsection{Prefix-based Training Augmentation}

\begin{algorithm}[!ht]
\caption{Prefix-based Training Augmentation}\label{alg:prefix}
\renewcommand{\algorithmicthen}{\textbf{then} \COMMENT{Truncated case}}
\renewcommand{\algorithmicelse}{\textbf{else} \COMMENT{Full-sequence case}}
\begin{algorithmic}[1]
\REQUIRE Training pair $(\mathbf{x}_{0:N-1}, \mathbf{y}_{0:M-1})$, probability $p_{\text{trunc}}$
\STATE $r \sim \text{Uniform}(0, 1)$
\IF{$r < p_{\text{trunc}}$}
    \STATE Sample $k \sim \text{Uniform}\{1, \dots, N-1\}$
    \STATE $\mathbf{x}' \gets x_{0:k-1}$ \COMMENT{No source-side \texttt{<EOS>}}
    \STATE $\mathbf{y}' \gets [y_{0:l-1}; \text{\texttt{<EOS>}}]$ \COMMENT{$y_{0:l-1}$ is the target aligned with $x_{0:k-1}$}
\ELSE
    \STATE $\mathbf{x}' \gets [\mathbf{x}_{0:N-1}; \text{\texttt{<EOS>}}]$
    \STATE $\mathbf{y}' \gets [\mathbf{y}_{0:M-1}; \text{\texttt{<EOS>}}]$
\ENDIF
\RETURN $(\mathbf{x}', \mathbf{y}')$
\end{algorithmic}
\end{algorithm}

The RTW policy requires the decoder to emit an \texttt{<EOS>} token to terminate each local normalization span. Standard training, however, typically conditions \texttt{<EOS>} prediction on a global source-side boundary. This creates a train-test mismatch during streaming inference, where local spans must terminate without observing a global \texttt{<EOS>} or future context. Simply appending an \texttt{<EOS>} token to source chunks during inference is unsuitable, as it would invalidate the incremental KV cache through self- and cross-attention and exacerbate distribution shifts.

To bridge this train-test discrepancy, we introduce \textbf{Prefix-based Training Augmentation} (Alg.~\ref{alg:prefix}), a computationally efficient method applied on-the-fly during training. We stochastically truncate training examples in an aligned manner with a probability $p_{\text{trunc}}$. In these instances, the source-side \texttt{<EOS>} is omitted, while the target-side prefix is explicitly terminated with an \texttt{<EOS>} token. 
This strategy compels the model to learn that any valid completion of a normalization span, even when occurring mid-sentence, must trigger termination based solely on the current partial encoder hidden states. By simulating these artificial boundaries, the model becomes robust to the truncated inputs encountered during streaming inference. Consequently, the decoder can reliably terminate generation without compromising the incremental KV cache, ensuring the system remains responsive to subsequent incoming chunks.

\subsection{Data Construction}
\label{sec:data_construction}
We construct our dataset from the \textit{binhvq}\footnote{\url{https://github.com/binhvq/news-corpus}} news corpus, a large-scale collection containing approximately 6 billion words across diverse domains. Using the proposed data generation pipeline, we produce 5M training sentence pairs, 50k validation pairs, and 50k test pairs. In addition to spoken-written text pairs, the dataset provides deterministic phrase-level alignments between spoken-form inputs and written-form targets, as well as token-level normalization boundary annotations. For span detection, each token is labeled using the IOB tagging scheme to identify segments requiring normalization. These spans encompass the entire affected word. For instance, a trailing punctuation mark is grouped within the preceding word's span. These annotations facilitate both prefix-based training augmentation and auxiliary tagging supervision. 
\\
\textbf{Categories}: We define a comprehensive set of ITN categories that reflect common phenomena in ASR outputs. These categories include punctuation restoration, capitalization, semiotic normalization, and phonetic normalization. For punctuation restoration, we consider four standard punctuation marks: comma, period, question mark, and exclamation mark \cite{ho25_interspeech}. Semiotic normalization covers entities such as numbers, dates, times, fractions, currencies, digit sequences, and electronic addresses, while phonetic normalization targets out-of-vocabulary (OOV) expressions, including foreign words, named entities, domain names, and letter sequences.
\\
\textbf{Pipeline}: The dataset is generated through an automated data generation pipeline that builds upon the data preparation framework introduced in \cite{nguyen2023adapitn}, with substantial extensions to support a broader range of ITN categories. Specifically, we define a comprehensive set of regular expressions (regex) to identify normalization regions in written-form text:
\begin{itemize}
    \item \textbf{Capitalization and punctuation restoration:} Spoken-form inputs are generated by removing all punctuation marks and converting the text to lowercase.
    \item \textbf{Semiotic normalization:} Each class (e.g., cardinal/ordinal numbers, decimals, currencies, time expressions, and factions) is identified using dedicated regex patterns. Since these expressions are detected in written form, rule-based transformations generate corresponding spoken-form candidates. Recognizing that a single written entity may have multiple valid spoken realizations, the pipeline randomly selects one candidate during data generation to reflect natural spoken variability. To ensure high quality, we manually inspect representative samples from each semiotic class to verify the correctness and coverage of the transformation rules.
    \item \textbf{Phonetic normalization:} We define phonetic phrases as tokens that are neither semiotic expressions nor present in the ASR vocabulary. This allows us to extract written-form phonetic entities, such as foreign named entities and domain-specific terms. We then perform manual verification of an automated grapheme-to-phoneme output to construct a dedicated phonetic dictionary comprising approximately 125k common foreign and OOV terms. This dictionary is subsequently used to map written-form phonetic phrases back to their spoken-form counterparts during dataset generation.

\end{itemize}
Through this regex-driven transformation process, the pipeline deterministically constructs phrase-level spoken-written alignments and assigns specific category labels to each normalization segment, providing precise supervision for both prefix-based training augmentation and the auxiliary tagging task for identifying normalization boundaries.

\section{Experimental Setup}
\label{sec:exp_setup}

\subsection{Datasets}

For training and evaluation, we utilize the dataset described in Section~\ref{sec:data_construction}. For hyperparameter tuning, we use a small subset of 500k samples from the training set and evaluate performance on the validation set. To better assess the system's real-world performance, we construct an additional test set by passing all 50k samples from the original test set through the Microsoft Edge TTS system\footnote{\url{https://github.com/rany2/edge-tts}} to synthesize speech. The generated audio is then processed by the ChunkFormer ASR model\footnote{\url{https://huggingface.co/khanhld/chunkformer-ctc-large-vie}}~\cite{le2025chunkformer} to produce ASR transcripts. The word error rate (WER) of this ASR output relative to the reference transcripts (spoken text) is $14.21\%$. This error rate reflects typical ASR artifacts, such as misrecognitions and phonetic ambiguities. These issues are especially pronounced because the dataset contains numerous OOV and foreign words, providing a challenging and realistic input stream. This ASR-generated text serves as the input for evaluating the robustness of our ITN model in a real-world deployment scenario, following the evaluation protocol in~\cite{gaur2023streaming}.

\subsection{Implementation Details}

\textbf{Training:} Models are trained for up to 20 epochs using 4 NVIDIA H100 (80GB) GPUs and are implemented via the Hugging Face Transformers framework\footnote{\url{https://github.com/huggingface/transformers}}. We employ the AdamW optimizer~\cite{loshchilov2018decoupled} with a peak learning rate of $10^{-4}$, a one-epoch linear warmup, and a cosine learning rate scheduler. Additional optimizer hyperparameters include $\beta_1 = 0.9$, $\beta_2 = 0.999$, weight decay $\lambda = 0.01$, and a global batch size of 512. Our streaming ITN models utilize dynamic context masking with variable chunk sizes $c \in \{1, 2, 3, 4, 5\}$ and right-context sizes $r \in \{0, 1, 2\}$ with $p_{\text{trunc}} = 0.8$ for prefix-based training augmentation. To benchmark against a streaming hybrid model, we retrain the architecture from \cite{ho25_interspeech} on our dataset and integrate it with our own weighted FST module. For a baseline streaming end-to-end approach, we train a multi-path wait-$k$ model~\cite{elbayad20_interspeech}. We also compare our approach against the AdapITN model~\cite{nguyen2023adapitn}, which is trained on our dataset with the biasing mechanism excluded to ensure a fair comparison.
\\
\textbf{Evaluation:} During inference, we employ greedy decoding. Following the methodology in~\cite{sunkara2021neural}, we evaluate performance using ITN Word Error Rate (I-WER) to measure normalization quality and Non-ITN Word Error Rate (NI-WER) to quantify unintended perturbations to the source text. Standard WER, the composite error of the entire process of converting spoken text to written text, is used for comprehensive performance assessment. Tagging performance is evaluated using the $F_1$-score. To simulate real-time inference for benchmarking, inputs are processed in discrete chunks. The reported metrics represent the average WER across an ensemble of configurations, where chunk sizes are sampled from $\{3, 4, 5\}$ tokens and right-context sizes are randomly selected from $\{1, 2\}$ tokens. For configurations utilizing the RTW policy, we set $L_{\text{max\_local}} = 64$. This evaluation strategy ensures that the model's robustness is accurately reflected across various streaming latencies, reflecting real-world usage scenarios.
\\
\textbf{Statistical Evaluation:} To ensure the reliability of our results, we employ a bootstrap method with 10,000 iterations to determine 95\% confidence intervals (CIs) for the metric of interest, leveraging the toolkit provided in~\cite{Confidence_Intervals}. Results are presented in the format $\textit{mean}_{[\textit{LB}, \textit{UB}]}$, where \textit{mean} represents the metric's value on the original test set and the interval $[\textit{LB}, \textit{UB}]$ denotes the lower and upper bounds of the 95\% confidence interval.

\section{Results and Analysis}
\label{sec:results}

\begin{table*}[!ht]
\caption{Evaluation of ITN performance with and without punctuation and capitalization (PnC). Parameters $c$ and $r$ denote chunk and right-context sizes during inference. All metrics are reported as percentages (\%). Model categories: $B$ (Non-streaming baselines), $H$ (Hybrid streaming models), and $S$ (Proposed end-to-end streaming variants).}
\label{tab:performance_comparison}
\setlength{\tabcolsep}{5.6pt} 
\centering
\begin{tabular}{cl ccc c ccc ccc}
\toprule
& & & & & & \multicolumn{3}{c}{\textbf{With PnC}} & \multicolumn{3}{c}{\textbf{Without PnC}} \\
\cmidrule(lr){7-9} \cmidrule(lr){10-12}
& \textbf{Model} & \textbf{Pretrained} & \textit{\textbf{c}} & \textit{\textbf{r}} & \textbf{F$_1$} & \textbf{I-WER} & \textbf{NI-WER} & \textbf{WER} & \textbf{I-WER} & \textbf{NI-WER} & \textbf{WER} \\
\midrule
\multirow{5}{*}{\rotatebox[origin=c]{90}{\textbf{Non-stream}}} 
& $B_0$ & $\times$ & $\infty$ & $\infty$ & -- & 14.20 & 1.51 & 4.61 & 8.28 & 0.54 & 1.11 \\
& $B_1$ & EnViT5 & $\infty$ & $\infty$ & -- & 11.69 & \textbf{1.30} & 3.84 & 5.17 & 0.39 & 0.74 \\
& $B_2$ & EnViT5 & $\infty$ & $\infty$ & 97.34 & \textbf{10.80} & 1.32 & \textbf{3.63} & \textbf{5.03} & \textbf{0.16} & \textbf{0.52} \\
& AdapITN~\cite{nguyen2023adapitn} & EnViBert & $\infty$ & $\infty$ & 96.65 & 156.88 & 2.50 & 40.08 & 14.58 & 0.40 & 1.45 \\
& $S_C$ (ours)& EnViT5 & \{3,4,5\} & \{0,1,2\} & 96.62 & 11.48 & 1.31 & 3.79 & 5.83 & 0.18 & 0.60 \\
\midrule
\multirow{8}{*}{\rotatebox[origin=c]{90}{\textbf{Stream}}}
& $H_1$~\cite{ho25_interspeech} & PhoBert & \{3,4,5\} & 0 & 86.30$^\dagger$ & 34.79 & 2.22 & 10.16 & 49.34 & 1.05 & 4.62 \\
& $H_2$~\cite{ho25_interspeech} & PhoBert & \{3,4,5\} & \{1,2\} & 90.21$^\dagger$ & 28.39 & 1.78 & 8.27 & 47.36 & 0.75 & 4.20 \\
& Wait-$k$~\cite{elbayad20_interspeech} & EnViT5 & \multicolumn{2}{c}{with $k = 9$} & -- & 19.40 & 3.00 & 6.99 & 16.79 & 2.68 & 3.72 \\
& $S_0$ (ours) & $\times$ & \{3,4,5\} & 0 & 95.09 & 21.39 & 1.56 & 6.39 & 10.54 & 1.03 & 1.73 \\
& $S_1$ (ours) & EnViT5 & \{3,4,5\} & 0 & 95.48 & 16.66 & 1.47 & 5.18 & 7.42 & 0.29 & 0.82 \\
& $S_2$ (ours) & EnViT5 & \{3,4,5\} & \{1,2\} & 97.19 & \textbf{12.91} & 1.36 & \textbf{4.18} & \textbf{6.24} & 0.19 & \textbf{0.64} \\
& $S_3$ (ours) & EnViT5 & \{3,4,5\} & 0 & 95.48 & 17.03 & 1.37 & 5.19 & 8.79 & 0.26 & 0.89 \\
& $S_4$ (ours) & EnViT5 & \{3,4,5\} & \{1,2\} & 97.19 & 13.22 & \textbf{1.31} & 4.21 & 7.30 & \textbf{0.16} & 0.69 \\
\bottomrule
\addlinespace[0.5ex]
\multicolumn{12}{l}{\footnotesize $^\dagger$ denotes the hybrid model's $F_1$ score, which is based on fine-grained multi-class classification.}
\end{tabular}
\end{table*}

\subsection{ITN Performance} 

In this section, we comprehensively evaluate the ITN performance of various model configurations, categorized into three groups:
\begin{itemize}
    \item \textbf{Non-streaming end-to-end baselines ($B$):} $B_0$ is an EnViT5 model trained from scratch, while $B_1$ utilizes pretrained weights. Both models are trained in a seq2seq manner using the $\mathcal{L}_{gen}$ loss function. $B_2$ further enhances $B_1$ with an additional auxiliary tagging loss, $\mathcal{L}_{tag}$. We also include AdapITN~\cite{nguyen2023adapitn} as a comparative baseline.
    \item \textbf{Hybrid streaming ($H$):} We benchmark against the hybrid architecture from~\cite{ho25_interspeech}, denoted as $H_1$ (with look-ahead $r=0$) and $H_2$ (with $r \in \{1, 2\}$).
    \item \textbf{Proposed end-to-end streaming ($S$):} $S_C$ is our causal-masked encoder model with non-streaming inference. $S_0$, $S_1$ and $S_2$ are our streaming variants using the RTW policy, whereas $S_3$ and $S_4$ represent the ``fully streaming'' versions utilizing decoder KV caching (as described in Section~\ref{sec:infer_kv_cache}). Note that $S_0$ is trained from scratch, while all other $S$ variants leverage pretrained weights. As an additional end-to-end streaming baseline, we benchmark against a multi-path wait-$k$ model~\cite{elbayad20_interspeech} decoded with $k = 9$.
\end{itemize}

Table~\ref{tab:performance_comparison} presents ITN performance across two settings: With PnC (punctuation, capitalization, and semiotic/phonetic normalization) and Without PnC (semiotic and phonetic normalization only).

\subsubsection{Non-Streaming Evaluation}

Leveraging the pretrained EnViT5 ($B_1$) yields significant gains over training from scratch ($B_0$), with relative WER improvements of $16.70_{[10.45, 26.15]}$\% and $33.33_{[5.33, 63.16]}$\% in the ``With PnC'' and ``Without PnC'' settings, respectively. Consistently lower I-WER and NI-WER scores highlight the practical benefits of pretrained text-to-text models for end-to-end ITN. By inheriting linguistic priors, the model better captures the global context and semantic alignment necessary to map informal speech to structured written forms. Incorporating the auxiliary tagging loss ($B_2$) further improves results, establishing a strong baseline with WERs of 3.63\% and 0.52\%.

Although AdapITN~\cite{nguyen2023adapitn} is an end-to-end system, it is not fine-tuned from a pretrained text-to-text model. Specifically, AdapITN utilizes encoder-only pretrained EnViBert~\cite{nguyen20d_interspeech} for both the encoder and decoder, with a tagging layer to isolate normalization regions. 
However, its decoder only processes isolated text spans, excluding the surrounding sentential context and making capitalization and punctuation recovery extremely difficult.
As a result, it struggles to generalize in the ``With PnC'' evaluation, yielding an I-WER of 156.88\% and a high overall WER of 40.08\%. Even in the ``Without PnC'' setting, which is consistent with its original evaluation, AdapITN's 1.45\% WER is $2.79_{[2.61, 3.30]}\times$ higher than our $B_2$ baseline. These results demonstrate the critical importance of incorporating surrounding context and leveraging a pretrained text-to-text model for complex ITN tasks.

\subsubsection{Streaming Evaluation}
Streaming performance follows a similar trend. $S_1$ (pretrained) outperforms $S_0$ (trained from scratch) with relative WER improvements of $18.94_{[10.92, 27.84]}$\% and $52.60_{[29.30, 67.08]}$\%. Our proposed $S_2$ variant achieves WERs of 4.18\% and 0.64\%, significantly outperforming the Wait-$k$ model~\cite{elbayad20_interspeech} by factors of $1.67_{[1.60, 1.75]}\times$ and $5.81_{[4.96, 6.73]}\times$, respectively. The failures of Wait-$k$ stem from forced generation: even after waiting $k$ tokens, the source input may still lag behind the target output, since a single target token can correspond to multiple source tokens. This situation occurs during both training and inference. Additionally, $S_2$ substantially surpasses the hybrid baseline $H_2$~\cite{ho25_interspeech} by $1.98_{[1.94, 2.01]}\times$ and $6.56_{[5.74, 7.40]}\times$. The high I-WERs observed for $H_2$ (28.39\% and 47.36\%) highlight the limitations of manually crafted FST rules for complex normalization, whereas our end-to-end approach learns these transformations directly from data.

Regarding the cost of streaming, the causal encoder with non-streaming inference ($S_C$) performs comparably to $B_2$, with marginal WER increases of $0.16_{[0.03, 0.27]}$\% and $0.08_{[-0.03, 0.18]}$\%. Under full streaming inference, $S_2$ remains robust: it incurs only an $8.85_{[-4.12, 16.77]}$\% relative WER increase over $B_1$ in the ``With PnC'' setting and improves performance by $13.51_{[-38.67, 50.62]}$\% in the ``Without PnC'' setting. This efficiency is attributed to our RTW policy and prefix-based training augmentation. Compared to $B_2$, $S_2$ shows relative WER increases of $15.15_{[10.29, 17.17]}$\% and $23.07_{[0.65, 47.15]}$\%, quantifying the inherent trade-off when future context is unavailable to resolve global dependencies. In terms of tagging accuracy, $S_2$ achieves a high $F_1$ score of 97.19\%, comparable to the $B_2$ baseline, demonstrating its reliable boundary detection capability for the RTW policy.

\subsection{Impact of KV Caching and Copy–Paste Strategy}

\begin{figure}[!th]
  \centering
  \includegraphics[width=\linewidth]{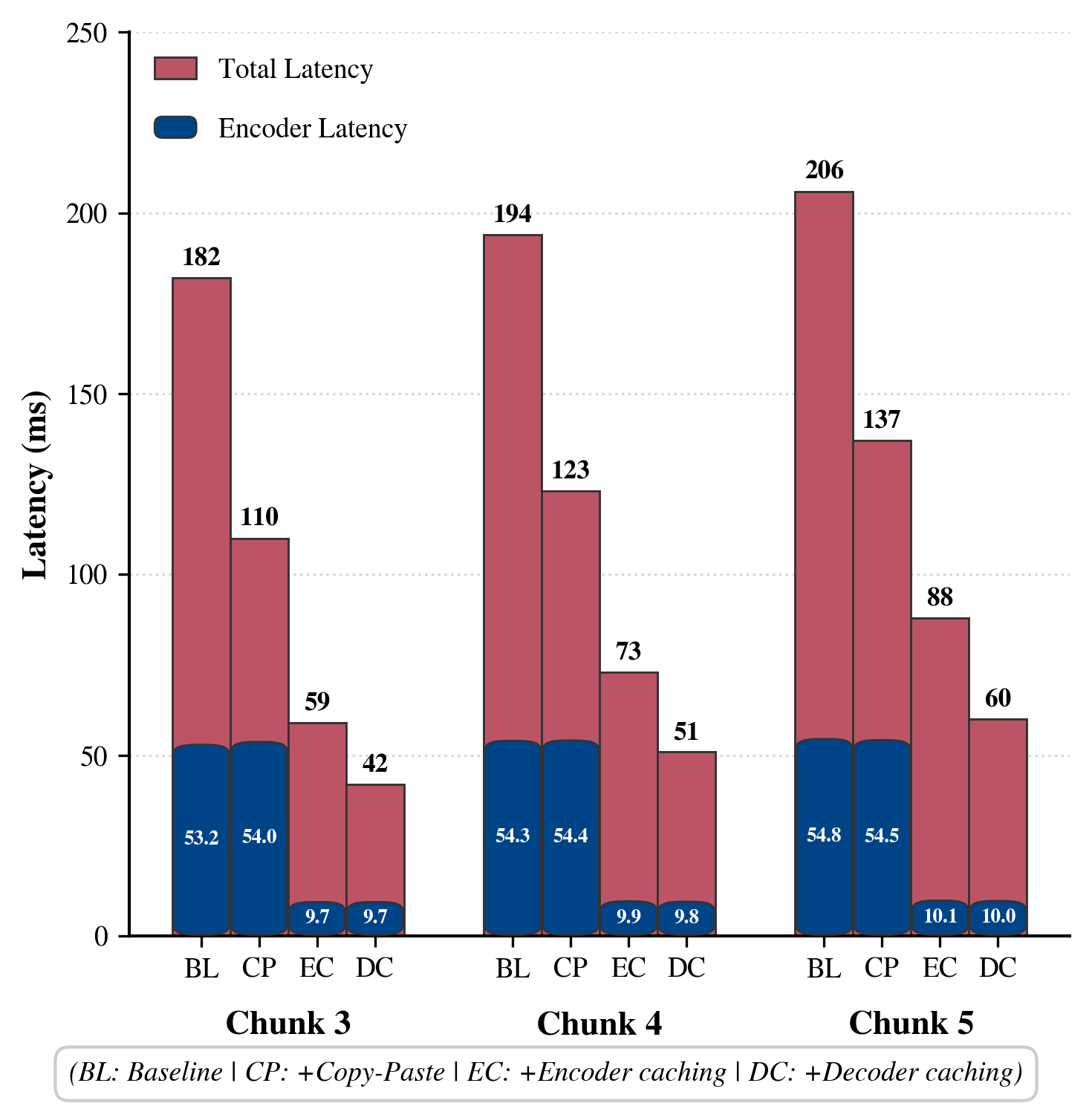}
  \caption{Per-chunk latency breakdown (Encoder vs. Total) for four streaming configurations across chunk sizes of 3, 4, and 5.}
  \label{fig:itn_latency}
\end{figure}

We benchmark per-chunk processing latency across configurations and chunk sizes with a fixed look-ahead $r=1$. To simulate high-concurrency production, we use a batch size of 100 with sequences of 100 - 200 words on an NVIDIA H100 GPU. Average encoder and total latencies are evaluated for four cumulative configurations, as shown in Fig.~\ref{fig:itn_latency}:

\begin{itemize}
  \item \textbf{BL (Baseline):} Standard streaming without optimizations, requiring full decoder processing for all tokens.
  \item \textbf{CP (+Copy–Paste):} Uses Copy–Paste strategy to bypass the decoder for $O$-tagged tokens, directly emit spoken-form text.
  \item \textbf{EC (+Encoder Caching):} Adds incremental encoder KV caching to CP (corresponding to $S_1$ and $S_2$ variants).
  \item \textbf{DC (+Decoder Caching):} The most optimized configuration, adding self- and cross-attention decoder KV caching to EC, representing the ``fully streaming'' $S_3$ and $S_4$ variants.
\end{itemize}

As illustrated in Fig.~\ref{fig:itn_latency}, the Copy–Paste strategy significantly reduces total latency and mitigates hallucinations by eliminating redundant decoder invocations for verbatim tokens. Encoder KV caching (EC) dramatically reduces encoder latency from 54~ms to 10~ms, while decoder KV caching (DC) further minimizes overhead by calculating updates only for new tokens, bypassing the initial batched forward pass. Combined, these optimizations achieve a $3.43\times$ speedup at $c=5$, where total latency decreases from 206~ms (BL) to 60~ms (DC). The total latency exhibits only marginal absolute increases across chunk sizes, offering flexible trade-offs between latency and context.

While the ``fully streaming'' DC approach introduces a minor training-inference mismatch, the performance impact is negligible. Table~\ref{tab:performance_comparison} shows that $S_4$ (DC) maintains nearly identical performance to $S_2$ (EC), with absolute WER increases of only $0.03_{[-0.05, 0.09]}$\% and $0.05_{[-0.03, 0.11]}$\% across settings. However, DC improves total per-chunk latency by 31.82\% relative to EC at $c=5$. We also evaluate the $95^{th}$ percentile ($p_{95}$) latency to assess worst-case scenarios requiring heavy normalization. The $p_{95}$ latency for DC is 165.5~ms, compared to 249.0~ms for EC. Given that a chunk size of 5 typically represents 500~ms of audio, our model ensures real-time processing even in the $95^{th}$ percentile, demonstrating its capacity to serve 100 concurrent users in large-scale deployments.

\subsection{Impact of Look-ahead}
As illustrated in Table~\ref{tab:performance_comparison}, incorporating look-ahead context ($S_2$) yields consistent performance gains across all primary metrics. Specifically, the $S_2$ model achieves relative WER reductions of $19.31_{[18.62, 19.89]}$\% and $21.95_{[17.95, 25.04]}$\% compared to the zero-look-ahead variant ($S_1$) in the ``With PnC'' and ``Without PnC'' settings, respectively. Similarly, the $S_4$ model achieves relative WER reductions of $18.88_{[18.50, 19.29]}$\% and $22.47_{[20.62, 23.99]}$\% compared to its zero-look-ahead counterpart ($S_3$). A comparable trend is observed in the hybrid streaming architecture, where $H_2$ outperforms $H_1$ with WER reductions of $18.60_{[18.33, 18.93]}$\% and $9.09_{[8.79, 9.42]}$\% in the corresponding settings. These findings underscore the critical role of look-ahead context during both training and inference.

\subsection{Real-world evaluation}

\begin{table}[!th]
  \caption{Real-world ITN performance on noisy ASR transcripts. All metrics are reported as percentages (\%).}
  \label{tab:realworld_eval}
  \centering
  \begin{tabular}{clccc}
    \toprule
    & \textbf{Model} & \textbf{I-WER} & \textbf{NI-WER} & \textbf{WER} \\
    \midrule
    \multirow{4}{*}[0.5ex]{\rotatebox[origin=c]{90}{\textbf{Non-stream}}} 
    & $B_0$ & 37.30 & 2.44 & 10.94 \\
    & $B_1$ & \textbf{34.09} & \textbf{2.24} & \textbf{10.01} \\
    & $B_2$ & 35.33 & 2.28 & 10.34 \\
    & $S_C$ & 36.18 & 2.27 & 10.54 \\
    \midrule
    \multirow{5}{*}{\rotatebox[origin=c]{90}{\textbf{Stream}}}
    & $H_2$ & 47.56 & 2.75 & 13.67 \\
    & Wait-$k$ & 41.16 & 2.48 & 11.91 \\
    & $S_0$ & 46.83 & 2.51 & 13.30 \\
    & $S_2$ & 37.55 & 2.29 & 10.89 \\
    & $S_4$ & \textbf{37.08} & \textbf{2.24} & \textbf{10.74} \\
  \bottomrule
  \end{tabular}
\end{table}

To assess performance in a practical deployment context, we evaluate our models on a test set of noisy ASR transcripts. Unlike the original test set, these inputs contain inherent recognition errors and acoustic artifacts typical of end-to-end ASR systems. The experimental results are summarized in Table~\ref{tab:realworld_eval}.

In the streaming scenario, our proposed $S_4$ variant achieves the highest performance, significantly outperforming the hybrid streaming baseline $H_2$ by  $21.43_{ [21.00, 21.95]}$\% and surpassing the Wait-$k$ by $9.82_{ [0.86, 19.46]}$\% in terms of relative WER reduction. When compared to the strongest offline baseline ($B_1$), $S_4$ maintains competitive performance, incurring a relative WER increase of only $7.29_{[6.85, 7.69]}$\%. Interestingly, on these noisy transcripts, the ``fully streaming'' model $S_4$ slightly surpasses the $S_2$ variant, with a relative improvement of $1.38_{[0.48, 2.73]}$\%. This suggests that the inclusion of decoder KV caching does not negatively impact robustness to ASR errors. Rather, it may provide slight stability in noisy streaming contexts.

\subsection{Impact of Truncation Probability}

\begin{figure}[!ht]
  \centering
  \includegraphics[width=\linewidth]{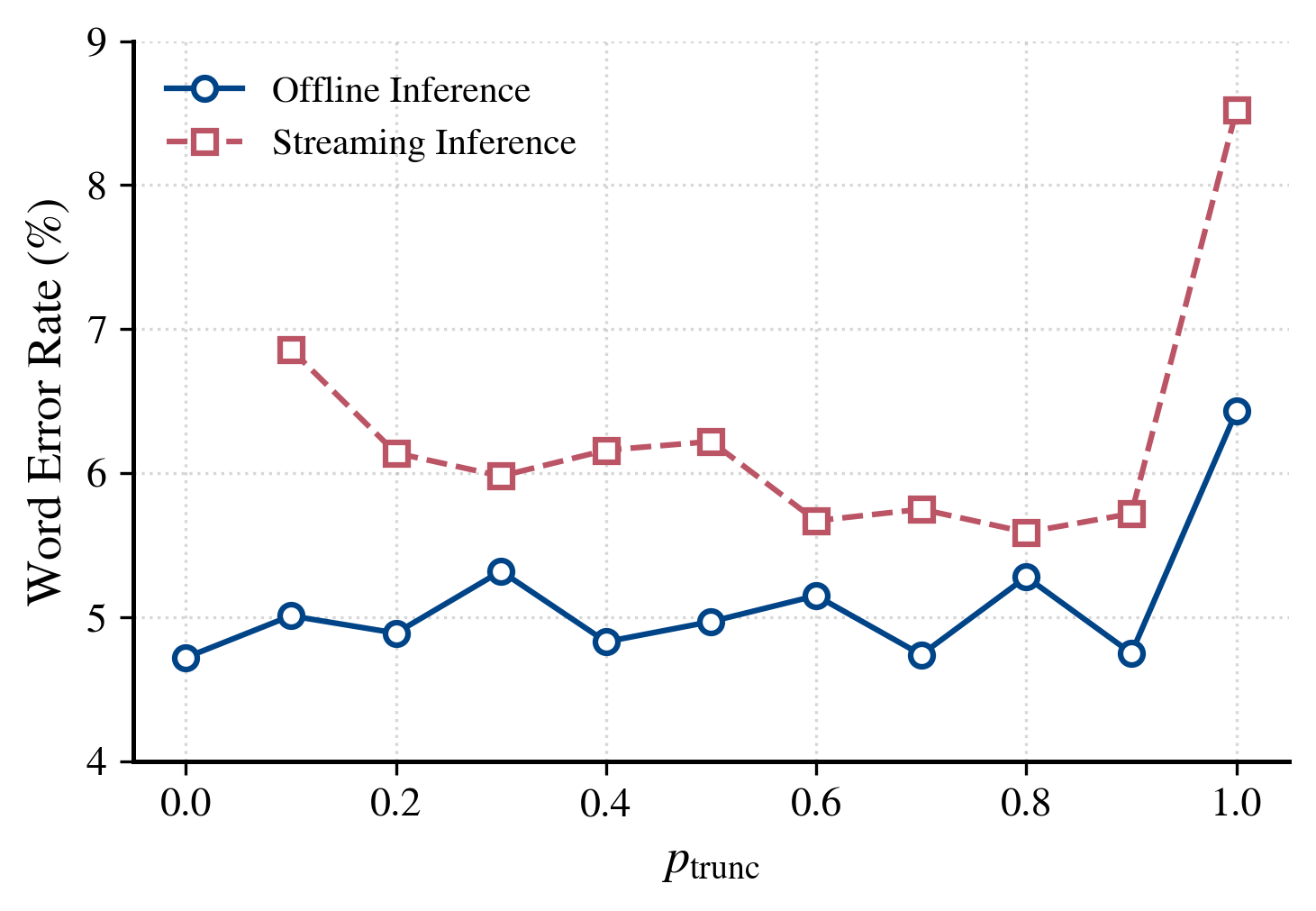}
  \caption{Impact of truncation probability $p_{\text{trunc}}$ on the Word Error Rate for both offline and streaming inference of our model.}
  \label{fig:wer_prob_trunc}
\end{figure}

To assess our prefix-based training augmentation, we evaluate performance on a validation subset across two modes: Streaming Inference (RTW policy) and Offline Inference (causal encoder). All experiments use $c{=}3$ and $r{=}1$. Figure~\ref{fig:wer_prob_trunc} illustrates the sensitivity of WER to the truncation probability $p_{\text{trunc}}$.

For offline inference, the model remains robust across $0.0 \le p_{\text{trunc}} \le 0.9$. While $p_{\text{trunc}} = 0$ yields the lowest WER by avoiding termination-related distribution shifts, performance stays stable as $p_{\text{trunc}}$ increases. A sharp degradation occurs only at $p_{\text{trunc}} = 1.0$, where the lack of natural sentence endings during training impairs generalization to full sequences.

In the streaming scenario, augmentation is essential. Without it ($p_{\text{trunc}} = 0$), the WER exceeds 100\% as the model fails to predict the local \texttt{<EOS>} tokens required by the RTW policy. Introducing prefix-based augmentation ($p_{\text{trunc}} > 0$) immediately stabilizes performance. As shown in Figure~\ref{fig:wer_prob_trunc}, increasing $p_{\text{trunc}}$ correlates with improved streaming accuracy, as the model learns to identify local boundaries from partial context.

The gap between offline and streaming modes narrows significantly for $p_{\text{trunc}} \in \{0.6, 0.7, 0.8, 0.9\}$. The optimal balance is achieved at $p_{\text{trunc}} = 0.8$, yielding the lowest streaming WER while maintaining competitive offline performance. The $p_{\text{trunc}} = 1.0$ case is also detrimental to streaming, suggesting that complete removal of full-sequence supervision harms long-range dependency modeling. These results confirm that prefix-based augmentation effectively bridges the training-inference discrepancy for streaming ITN.

\subsection{Generalizability Across Architectures and Languages}

To demonstrate the versatility of our framework, we evaluate it across alternative text-to-text architectures, pretraining objectives, and languages. For Vietnamese, we adopt the pretrained BartPho model\footnote{\url{https://huggingface.co/vinai/bartpho-syllable-base}} \cite{tran22b_interspeech} as the backbone, which was pretrained using a denoising autoencoder objective on Vietnamese text. For English, we leverage a pretrained T5 model\footnote{\url{https://huggingface.co/google-t5/t5-base}} to assess cross-lingual compatibility, with training and evaluation conducted on the English subset of the standard Google TN dataset~\cite{sproat17_interspeech}.


\begin{table}[!th]
\caption{Generalizability of ITN performance across architectures and languages. Values represent average WER (\%) across $c \in \{3,4,5\}$ and $r \in \{0,1,2\}$. Vietnamese evaluation uses ASR-L (ground-truth spoken text) and ASR-O (noisy ASR transcripts); English evaluation uses With and Without PnC.}
\label{tab:architecture_generalizability}
\centering
\renewcommand{\arraystretch}{0.93} 
\setlength{\tabcolsep}{3.2pt}
\begin{tabular}{llcccc}
\toprule
\textbf{Lang.} & \textbf{Backbone} & \textbf{Params} & \textbf{Stream} & \multicolumn{2}{c}{\textbf{WER (\%)}} \\
\midrule
 & & & & \textbf{ASR-L} & \textbf{ASR-O} \\
\cmidrule(lr){5-6}
\multirow{4}{*}{Vi} & \multirow{2}{*}{BartPho} & \multirow{2}{*}{132M} & $\times$  & 4.61 & 11.07 \\
                    &                          &                       & $\checkmark$ & 6.31 & 12.84 \\
\cmidrule(lr){2-6}
                    & \multirow{2}{*}{EnViT5}  & \multirow{2}{*}{275M} & $\times$  & 3.84 & 10.01 \\
                    &                          &                       & $\checkmark$ & 4.54 & 11.02 \\
\midrule
 & & & & \textbf{W/ PnC} & \textbf{W/o PnC} \\
\cmidrule(lr){5-6}
\multirow{2}{*}{En} & \multirow{2}{*}{T5}      & \multirow{2}{*}{223M} & $\times$  & 6.48 & 1.33 \\
                    &                          &                       & $\checkmark$ & 8.17 & 1.84 \\
\bottomrule
\end{tabular}
\end{table}

We evaluate streaming models under the RTW policy and non-streaming models using standard decoding. Table~\ref{tab:architecture_generalizability} reports the average WER across all configurations ($c \in \{3,4,5\}$ and $r \in \{0,1,2\}$). The results show that our framework consistently narrows the performance gap between streaming and non-streaming settings across architectures and languages. While the larger EnViT5 backbone delivers the best absolute performance, the more compact BartPho model demonstrates comparable robustness under streaming constraints and noisy ASR inputs. A similar trend is observed in English, where the T5 backbone also achieves strong and stable performance. Together, these results suggest that the proposed method is architecture-agnostic and language-agnostic, maintaining effectiveness across different model scales and linguistic contexts.


\subsection{Limitations and Future Work}

Despite the efficiency gains of our streaming ITN framework, several limitations offer avenues for future research:

\begin{itemize}
    \item \textbf{Phonetic Normalization and Failure Modes:} While semiotic classes are handled effectively, qualitative error analysis reveals that performance on OOV expressions and foreign entities remains challenging. A common failure mode occurs when the tagger successfully identifies a phonetic span, but the decoder hallucinates an orthographically plausible but factually incorrect spelling for a rare foreign entity (e.g., generating ``Marid'' instead of ``Madrid'' due to phonetic ambiguity in the ASR output). Future work will investigate biasing mechanisms to integrate phonetic dictionaries directly into the decoding process to improve I-WER without disrupting the incremental KV cache or increasing latency.
    
    \item \textbf{Early Emission Strategy:} The current RTW policy waits for a complete $B-I$ span before emission, which may introduce perceived lag for long sequences. We aim to explore predictive decoding strategies that allow for partial output as soon as sufficient linguistic context is captured, rather than waiting for span completion.

    \item \textbf{Edge Deployment Optimization:} With 275M parameters, the model may limit deployment on resource-constrained devices. Future work will explore compression methods, such as knowledge distillation and quantization, to reduce memory footprint and improve throughput on low-power hardware.
\end{itemize}
\section{Conclusion}
\label{sec:conclusion}

This paper introduces an efficient end-to-end streaming ITN framework leveraging pretrained text-to-text models. By integrating a chunk-based causal encoder with an RTW decoding policy, we adapt non-streaming architectures for real-time inference. To bridge the train-test discrepancy, we propose a prefix-based training augmentation strategy that ensures reliable local termination. Combined with KV caching and a Copy–Paste mechanism, our system meets production-grade latency requirements without sacrificing accuracy. Experimental results demonstrate that our approach significantly outperforms traditional hybrid streaming methods, achieving performance comparable to offline baselines. Finally, evaluations across architectures and languages confirm that the framework is scalable, backbone-agnostic, and language-agnostic, offering a versatile solution for complex normalization tasks.

\section{Acknowledgments}
We thank UNEY for sponsoring this research and providing the computational resources essential for model training and evaluation.

\section{Generative AI Use Disclosure}
Generative AI tools were used solely for editing and language polishing of the manuscript. They were not used to generate scientific content, experimental designs, results, analyses, or conclusions. All authors are responsible for the full content of this paper and consent to its submission.

\bibliographystyle{IEEEtran}
\bibliography{mybib}

\end{document}